# Five Years of COVID-19 Discourse on Instagram: A Labeled Instagram Dataset of Over Half a Million Posts for Multilingual Sentiment Analysis

Nirmalya Thakur
Department of Electrical Engineering and Computer Science
South Dakota School of Mines and Technology
Rapid City, SD 57701, USA
nirmalya.thakur@sdsmt.edu

***Abstract*—The outbreak of COVID-19 served as a catalyst for content creation and dissemination on social media platforms, as such platforms serve as virtual communities where people can connect with one another seamlessly. While there have been several works related to the mining and analysis of COVID-19-related posts on social media platforms such as Twitter (or X), YouTube, Facebook, and TikTok, there is still limited research that focuses on the public discourse about COVID-19 on Instagram. Furthermore, the prior works in this field have only focused on the development and analysis of datasets of Instagram posts published during the first few months of the outbreak. The work presented in this paper aims to address this research gap and makes three scientific contributions to this field. First, it presents a multilingual dataset of 500,153 Instagram posts about COVID-19 published between January 2020 and September 2024. This dataset, available at https://dx.doi.org/10.21227/d46p-v480, contains Instagram posts in 161 different languages, and these posts contain 535,021 distinct hashtags. After the development of this dataset, multilingual sentiment analysis was performed, which involved classifying each post as positive, negative, or neutral. The results of sentiment analysis are presented as a separate attribute in this dataset. Second, the paper presents the results of performing sentiment analysis per year from 2020 to 2024. The findings revealed the trends in sentiment related to COVID-19 on Instagram since the beginning of the pandemic. For instance, between 2020 and 2024, the sentiment trends show a notable shift, with positive sentiment decreasing from 38.35% to 28.69%, while neutral sentiment increasing from 44.19% to 58.34%. Finally, the paper also presents findings of language-specific sentiment analysis. This analysis highlighted similar and contrasting trends of sentiment across posts published in different languages on Instagram. For instance, out of all English posts, 49.68% were positive, 14.84% were negative, and 35.48% were neutral. In contrast, among Hindi posts, 4.40% were positive, 57.04% were negative, and 38.56% were neutral, reflecting distinct differences in the sentiment distribution between these two languages.**

## I. Introduction

The first few cases of the COVID-19 pandemic, caused by the SARS-CoV-2 virus, were detected in December 2019 in Wuhan, China. Since then, the virus rapidly spread to all parts of the world, leading to an unprecedented number of cases and deaths, the likes of which humanity has not experienced in centuries [1]. On March 11, 2020, the World Health Organization (WHO) declared COVID-19 an emergency [2]. As of September 15, 2024, there have been 776,281,230 cases and 7,065,880 deaths due to COVID-19 on a global scale [3]. COVID-19 led to major disruptions in global economies. During the first few months of the outbreak, the imposition of lockdowns in different parts of the world led to the interruption of supply chains, unemployment, disruption of education, reduced physical activity, and increased mental health issues [4-7]. In addition to this, the healthcare sector was significantly overwhelmed due to an increase in demand for services and supplies [8,9].

In the last decade and a half, social media platforms have emerged as invaluable sources for seeking and sharing information during virus outbreaks. The patterns of content creation and dissemination on social media can be interpreted and analyzed using concepts of Natural Language Processing, Machine Learning, and Data Science to understand web behavior, infer public sentiment, identify misinformation, evaluate the effectiveness of public health campaigns, track mental health issues, detect real-time crisis communication, track supply shortages, and identify trends in preventive behavior, related to virus outbreaks [10-13]. As a result, the mining and analysis of social media data to conduct syndromic surveillance, with a specific focus on public health, has attracted the attention of researchers from different disciplines in the recent past [14-17].

During the COVID-19 pandemic, the usage of social media platforms increased tremendously. Internet use increased by 50-70%, and social media platforms such as Instagram, Twitter, Facebook, and TikTok saw increased engagement as people sought both entertainment and social interaction during periods of isolation [18]. A study focusing on adolescents in the United States reported an average screen time of 7.7 hours per day during the pandemic, most of which was dedicated to social media use [19]. This increase in social media usage not only underscored the consistent reliance of the global population on digital platforms for maintaining social connections but also resulted in the generation of tremendous amounts of Big Data. This Big Data served as a rich resource for researchers from different disciplines to investigate a wide range of research questions related to the patterns of content creation and dissemination about COVID-19 on social media. The significant increase in the usage of social media platforms also facilitated the rapid spread of both reliable information and misinformation related to the pandemic [20,21]. Instagram, a globally popular social media platform with over 2 billion users [22], has become even more popular since the outbreak of COVID-19. Instagram's emphasis on imagery and short videos allowed users to convey complex emotions and narratives succinctly, fostering a sense of community and shared experience during the pandemic [23,24]. This increasing popularity of Instagram since the beginning of the COVID-19 outbreak, its diverse user base, and global reach highlight its potential as a rich resource for analyzing public sentiment and discourse related to COVID-19 as well as for investigating language diversity and cultural trends that have emerged on Instagram in this context. Despite the global popularity of Instagram, there is still limited research related to the mining and analysis of the public discourse about COVID-19 on Instagram.

In the past couple of years or so, there have been multiple works that focused on the development and analysis of datasets of social media posts about COVID-19 in different languages such as English [25-28], Bengali [29,30], Spanish [31, 32], Turkish [33-35], Arabic [36-38], Hindi [39-41], Indonesian [42-44], and Italian [45-47], just to name a few. However, most of these works have not focused on the mining and analysis of posts about COVID-19 on Instagram. Furthermore, these works have primarily focused on a specific language and analyzed posts published during the first few months of the outbreak. While multiple multilingual datasets of social media posts about COVID-19 do exist (for example: [48-54]), those datasets contain Tweets (or posts on X) about COVID-19. A prior work by Zarei et al. [55] involved the development of a dataset of Instagram posts about COVID-19. However, that work contains posts published only during the first few months of the pandemic, from January 5, 2020 to March 30, 2020. In addition to the above, no prior work in this field has focused on the development of an Instagram dataset, which is labeled for sentiment analysis. A multilingual dataset of Instagram posts about COVID-19, labeled for sentiment analysis, is expected to serve as a resource for the investigation of a wide range of research questions, such as:

(1) How does sentiment toward COVID-19 vary across different languages?
(2) How has public sentiment toward COVID-19 evolved from 2020 to the present?
(3) How do cultural differences affect social media discourse about COVID-19 across various languages?
(4) How has COVID-19 impacted mental health, as reflected in social media posts across different languages?
(5) What forms of vaccine hesitancy or support appear in different languages?
(6) How did seasonal or significant COVID-19 milestones (e.g., vaccine rollouts, lockdowns, emergence of variants) influence sentiment across different languages on Instagram?
(7) How did geopolitical events influence public sentiment about COVID-19?
(8) What role does social media discourse play in shaping public behavior toward COVID-19 in different linguistic communities?
(9) What are the trends in sentiment toward COVID-19 between less common languages and widely spoken languages (e.g., English, Spanish)?
(10) How do specific COVID-19 hashtags correlate with positive, negative, or neutral sentiments in Instagram posts across different languages?

In addition to the above, such a dataset will also be helpful for the training and testing of machine learning models for sentiment analysis of social media posts related to COVID-19. The work of this paper aims to address these research gaps. It presents a multilingual dataset of more than half a million posts about COVID-19 on Instagram, labeled for sentiment analysis. These posts are available in 161 different languages and were published on Instagram between January 2020 and September 2024. The paper also presents the findings of performing sentiment analysis per year from 2020 to 2024, as well as the findings of performing language-specific sentiment analysis.

The rest of this paper is organized as follows. Section II presents a review of recent research works in this field and discusses the research gaps in detail. The methodology is explained in Section III, which is followed by the results in Section IV. Section V concludes the paper, which is followed by the references.

## II. Literature Review

The COVID-19 pandemic led to a tremendous increase in the use of social media platforms, including Instagram. There have been multiple studies that have focused on mining and analysis of the patterns of content creation and dissemination about COVID-19 on Instagram. Priadana et al. [56] analyzed 10,403 Instagram posts containing #wabahcorona published between February 28, 2020, and May 18, 2020. Their study highlighted the role of hashtags in creating COVID-19-related content on Instagram. Rogowska et al. [57] recruited 954 students between the ages of 19 and 42 for their study. The results showed that the prevalence of Instagram addiction, loneliness, and dissatisfaction with life was 17.19%, 75%, and 40.15%, respectively. The work of Lee et al. [58] focused on studying the usage of #slowfashionaustralia in Instagram posts about COVID-19. The work of Quinn et al. [59] focused on misinformation analysis on Instagram in the context of COVID-19. They analyzed posts containing #hoax, #governmentlies, and #plandemic published between April 21, 2020, and April 30, 2020. Their study resulted in the identification of multiple themes related to general mistrust and conspiracy theories in the context of the public discourse on Instagram about COVID-19 during that time. Rajan et al. [60] explored how right-wing nationalist movements used Instagram to spread Islamophobia during the pandemic to illustrate how social media platforms became vehicles for political and social tensions during the crisis. Basch et al. [61] collected Instagram posts containing #momjuice and #winemom with a specific focus on studying alcohol-related content posted by mothers on Instagram during COVID-19. The work of Er et al. [62] involved studying Instagram posts by parents in the context of COVID-19 that included #korona (corona) and #evdekal (stay at home). The authors analyzed 401 Instagram posts published by parents about their children between April 18, 2020, and April 30, 2020.

Sui et al. [63] studied fitness videos on YouTube and Instagram to infer the levels of engagement with a specific focus on views, likes, and comments. Their work involved studying videos published between March 11, 2020, and June 30, 2020. The findings of their work showed that for every channel, the peak engagement was associated with the initial video, and thereafter, the engagement gradually declined. The work of Tuomi et al. [64] involved using a mixed methods approach to investigate the patterns of Instagram usage by high-profile Finnish restaurants during the COVID-19-related national lockdown. They studied 1119 Instagram posts published by 45 restaurants and interviewed restaurant managers. The findings of their work showed that the number of likes on Instagram posts by these restaurants during the lockdown due to COVID-19 stayed relatively similar to the number of likes before the lockdown. However, there was a significant increase in the number of comments on Instagram posts during the lockdown.

Wati et al. [65] aimed to infer the effectiveness of Instagram as a promotional tool during COVID-19. They analyzed the Instagram posts of a popular restaurant, *FOS Food Mojokerto*, in their study. Dušek et al. [66] studied the variations in Instagram usage in university students before and during COVID-19. A similar study that focused on analyzing Instagram usage during COVID-19 was performed by Dou et al. [67]. Aufa et al. [68] examined the engagement rates of Instagram accounts of different hospitals and found that health-related posts saw a significant increase in engagement during COVID-19. Lucibello et al. [69] studied 668 Instagram posts containing #quarantine15 to interpret the patterns of content creation and dissemination during COVID-19 with a specific focus on body image-related matters. The results showed that the posts containing human figures showcased individuals who were perceived as underweight, white, and women, and approximately one-third of these images were considered objectifying. Amanatidis et al. [70] studied Instagram posts published by three companies involved in COVID-19 vaccine research. The findings revealed a neutral to negative sentiment, with highly polarized user post distributions. Niknam et al. [71] studied 1612 Instagram posts about COVID-19 published between February 19, 2020, and April 3, 2020. They identified 23 themes of conversations in these posts.

In summary, even though several works in this field have focused on the investigation of different research questions with a specific focus on mining and analyzing Instagram posts about COVID-19, multiple research gaps still remain. First, there is a need for a multiyear dataset of Instagram posts about COVID-19. Most studies have been limited to short-term analysis, often focusing on the early stages of the pandemic. A multiyear dataset would provide valuable insights into how public sentiment, behavior, and engagement have evolved over time, especially as the pandemic progressed through different phases, such as the introduction of vaccines, the emergence of new variants, and shifting public health measures. Second, there is a need for a multilingual dataset. Much of the existing research has focused on English-language posts, which limits the generalizability of findings to non-English-speaking populations. Given the global nature of the pandemic, a dataset that includes posts in multiple languages would allow for a more comprehensive understanding of how COVID-19 was experienced across different cultural and linguistic contexts. Finally, a labeled dataset is needed for sentiment analysis. While sentiment analysis has been used in COVID-19-related Instagram studies, the absence of labeled datasets has limited the usage and application of supervised learning models in this context. A labeled dataset is expected to contribute to the development and optimization of supervised learning models for sentiment analysis of Instagram posts about COVID-19. The work presented in this paper aims to address these research gaps. The step-by-step methodology that was followed for the development of the dataset and for performing the data analysis is presented in Section III.

## III. Methodology

The dataset was developed by mining Instagram posts that contained at least one hashtag related to COVID-19, published on Instagram between January 21, 2020, and September 23, 2024. In this context, September 23, 2024, was the most recent date at the time of development of this dataset. To perform the data mining process, prior works in this field and trending hashtags related to COVID-19 were reviewed, and a list of hashtags related to COVID-19 was developed. These hashtags are shown in Table 1. A program was written in Python 3.11.5 for the development of this dataset and the data mining of the relevant Instagram posts, i.e., the posts that contained at least one hashtag from Table 1 and were published between January 21, 2020, and September 23, 2024, was performed by connecting to the Instagram API. All the Instagram posts that were collected during this data mining process were publicly available on Instagram and did not require a user to log in to Instagram to view the same (at the time of writing of this paper).

Table 1: Hashtags related to COVID-19 which were used for Data Mining

| | |
|---|---|
| #covid | #CoronavirusLockdown |
| #covid_19 | #postcovid19 |
| #coronavirus | #freecovidtesting |
| #covid19 | #covid19outbreak |
| #covidtesting | #Unite2FightCorona |
| #CoronaVirusUpdates | #covid2020 |
| #corona | #covidkindness |
| #covidpandemic | #CoronaUpdate |
| #coronavirusoutbreak | #coronavirusmemes |
| #covidvaccine | #longcovidsymptoms |
| #covidupdate | #covidmemes |
| #covidrelief | #CoronaVaccine |
| #covidheroes | #postcovidsyndrome |
| #covidtest | #coronaupdates |
| #covidvacccine | #coronapandemic |
| #IndiaFightsCorona | #CoronaLockdown |
| #CoronavirusPandemic | #covidsafe |
| #coronavirusindia | #covidhelp |
| #coronavirusawareness | #treatlongcovid |
| #covidsupport | #longcovidwarrior |
| #coronavirusvaccine | #covidresponse |
| #covidindia | #covidrescue |
| #coronavirusnews | #covidlife |
| #covid19testing | #alert_covid19 |
| #covid19india | #keralamodelcovid19prevention |
| #coronavirus2020 | #covidrecovery |
| #COVID2019 | #covidfree |
| #covidnews | #covidtimes |
| #coronavirusupdate | #CoronaUpdatesInIndia |
| #longcovid | #coronawarriors |
| #coronamemes | #covidrestrictions |
| #covidpositive | #coronanews |
| #covid19news | #CoronavirusOutbreakindia |
| #covidupdates | #coronaviruschina |
| #coronaindia | #coronaviruses |
| #covidwarriors | #covidbooster |
| #postcovid | #covidtestkit |
| #coronavirusinindia | #fightagainstcorona |
| #covid19vaccine | #covidusa |
| #CoronaAlert | #coronaoutbreak |
| #covidtravel | #covid19test |
| #covid19pandemic | #covidcases |
| #covidresources | #covidtestingsites |
| #coronavírus | #coronavirusprevention |

After performing data mining, the Google Translate API [72] was used to perform language detection. A program was written in Python, and the Google Translate API was called by this program. As this dataset contains more than 500,000 posts, the associated Google Cloud account was set up with a payment method such that the costs for the API call were automatically paid from this payment method every time an invoice was generated in the Google Cloud account, and the program did not crash or terminate on account of not being able to make subsequent calls to the Google Translate API due to any pending invoice(s). The output of this program resulted in the addition of two new columns to the dataset file. One of these columns represented the language code (for example, "en"), and the other column represented the full form of the language code (for example, "English"). To obtain the data for both of these columns, separate calls had to be made to the Google Translate API. The cost of using the Google Translate API for this purpose was about $4000.00. The next step was data preprocessing, which was also performed by writing a program in Python 3.11.5. First, all the posts were converted to lowercase, and hashtags, user mentions, and numbers were removed using regular expressions. Then, a function was written to remove any irrelevant symbols, punctuation marks, and emojis that do not convey any emotion. To develop this function, a list of emojis was developed, which are usually used in social media posts to convey different types of emotions (as shown in Table 2). This function retained only these emojis in the Instagram posts and deleted all the other emojis (a few examples of emojis that were deleted during this process are 🔬, ✈️, 📆, 🏃‍♀️, ✅, 📞, 🦠, 🍁, 👉🏼, 🛡️, ☝️, 📚, 📸, 🏥, 🤴, 📊, 📈, 💰, and 💫).

The next step for data preprocessing was stop word removal. This was performed using the stopwordsiso package [73] by passing the language code for each post as input. This package was specifically used because it supports the removal of stop words in different languages. After the data preprocessing was completed, sentiment analysis was performed. For posts published in English, VADER (Valence Aware Dictionary and sEntiment Reasoner) was applied [74] as it is specifically designed for analyzing social media texts. VADER computed the compound sentiment score for every English post, and based on the compound sentiment score, it classified a post as positive, negative, or neutral. For non-English posts, the sentiment was analyzed using the twitter-xlm-roberta-base-sentiment model from HuggingFace's transformers library [75]. This multilingual model is a fine-tuned version of RoBERTa, trained to classify text into positive, neutral, or negative sentiments across different languages. This model required posts longer than 945 characters to be truncated due to input size constraints. For VADER, this limitation did not exist. The output of performing this multilingual sentiment analysis, i.e., the sentiment label for every post, was added as an attribute in the dataset. After the development of this dataset, another program was written in Python 3.11.5 to identify the annual variations of sentiment and to perform language-specific sentiment analysis. The results obtained from these programs are discussed in Section 4.

Table 2: List of Emojis that were retained during Data Preprocessing

| Emoji Symbol | Emoji Description | Emoji Symbol | Emoji Description |
|---|---|---|---|
| 😀 | grinning face | 🥵 | hot face |
| 😃 | grinning face with big eyes | 🥶 | cold face |
| 😄 | grinning face with smiling eyes | 😱 | face screaming in fear |
| 😁 | beaming face with smiling eyes | 😨 | fearful face |
| 😆 | grinning squinting face | 😰 | anxious face with sweat |
| 😅 | grinning face with sweat | 😥 | sad but relieved face |
| 😂 | face with tears of joy | 😓 | downcast face with sweat |
| 🤣 | rolling on the floor laughing | 🤔 | thinking face |
| 😊 | smiling face with smiling eyes | 🤗 | hugging face |
| 😇 | smiling face with halo | 🤭 | face with hand over mouth |
| 🙂 | slightly smiling face | 🤫 | shushing face |
| 🙃 | upside-down face | 😐 | neutral face |
| 😉 | winking face | 😑 | expressionless face |
| 😍 | smiling face with heart eyes | 😶 | face without mouth |
| 😘 | face blowing a kiss | 😯 | hushed face |
| 😗 | kissing face | 😦 | frowning face with open mouth |
| 😙 | kissing face with smiling eyes | 😧 | anguished face |
| 😚 | kissing face with closed eyes | 😮 | face with open mouth |
| 😋 | face savoring food | 😲 | astonished face |
| 😛 | face with tongue | 😴 | sleeping face |
| 😝 | squinting face with tongue | 🤤 | drooling face |
| 😜 | winking face with tongue | 😪 | sleepy face |
| 🤪 | zany face | 🤐 | zipper-mouth face |
| 🤨 | face with raised eyebrow | 🥴 | woozy face |
| 🧐 | face with monocle | 😷 | face with medical mask |
| 🤓 | nerd face | 🤒 | face with thermometer |
| 😎 | smiling face with sunglasses | 🤕 | face with head-bandage |
| 😏 | smirking face | 🤢 | nauseated face |
| 😒 | unamused face | 🤮 | face vomiting |
| 😞 | disappointed face | 🤧 | sneezing face |
| 😔 | pensive face | 😈 | smiling face with horns |
| 😟 | worried face | 👿 | angry face with horns |
| 😕 | confused face | 💀 | skull |
| 🙁 | slightly frowning face | ☠️ | skull and crossbones |
| ☹️ | frowning face | 👻 | ghost |
| 😣 | persevering face | 👽 | alien |
| 😖 | confounded face | 🤖 | robot |
| 😫 | tired face | 🎃 | jack-o-lantern |
| 😩 | weary face | 😺 | grinning cat |
| 😢 | crying face | 😸 | grinning cat with smiling eyes |
| 😭 | loudly crying face | 😹 | cat with tears of joy |
| 😤 | face with steam from nose | 😻 | smiling cat with heart-eyes |
| 😠 | angry face | 😼 | cat with wry smile |
| 😡 | pouting face | 😽 | kissing cat |
| 🤬 | face with symbols on mouth | 🙀 | weary cat |
| 🤯 | exploding head | 😿 | crying cat |
| 😳 | flushed face | 😾 | pouting cat |
| 😵 | dizzy face | | |

## IV. RESULTS AND DISCUSSIONS

The dataset that was developed is available on IEEE Dataport at https://dx.doi.org/10.21227/d46p-v480. It contains 500,153 Instagram posts about COVID-19 published between January 21, 2020, and September 23, 2024. There are a total of 535,021 distinct hashtags present in this dataset. Table 3 presents a data description of this dataset. As stated in Table 3, this dataset presents the IDs of these posts instead of the URLs of these posts to prevent direct identification of the Instagram users who published these posts about COVID-19. For any post on Instagram, if the Post ID is known, it can be substituted in "PostIDhere" in the generic representation of an Instagram URL: https://www.instagram.com/p/PostIDhere/, to obtain the complete URL of that post.

The Instagram posts in this dataset are present in 161 different languages out of which the top 10 languages in terms of frequency are English (343041 posts), Spanish (30220 posts), Hindi (15832 posts), Portuguese (15779 posts), Indonesian (11491 posts), Tamil (9592 posts), Arabic (9416 posts), German (7822 posts), Italian (5162 posts), and Turkish (4632 posts). Figure 1 shows a word cloud of all the languages present in this dataset. The results of performing sentiment analysis to identify the trends of sentiment related to COVID-19 per year between 2020 and 2024 are shown in Figures 2 to 6.

Table 3. Data Description of this dataset

| **Attribute Name** | **Attribute Description** |
|---|---|
| PostID | Unique ID of each Instagram post |
| Post Description | Complete description of each post in the language in which it was originally published |
| Date | Date of publication of each post in MM/DD/YYYY format |
| Language Code | The language code of each post (for example, "en") |
| Full Language | The full form of the language of each post (for example, "English") |
| Sentiment | Results of sentiment analysis (using the preprocessed version of each post) where each post was classified as positive, negative, or neutral |

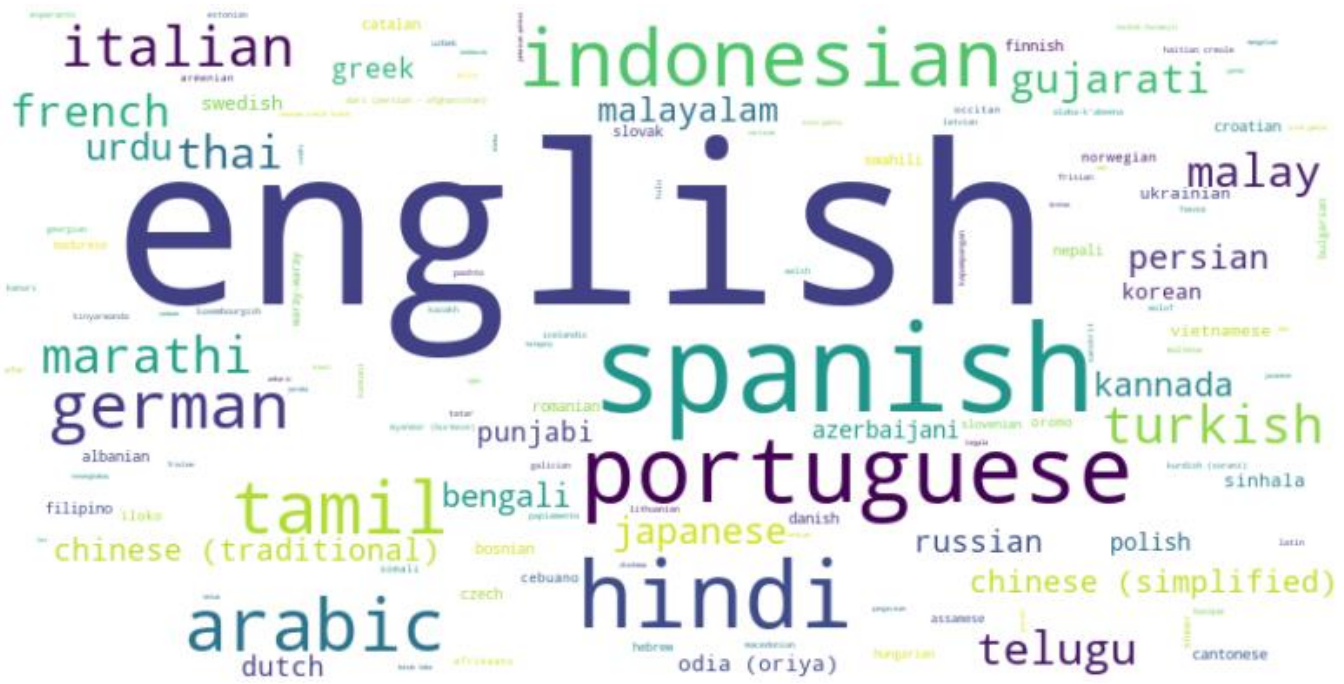

Figure 1. A word cloud representation of all the languages present in this dataset

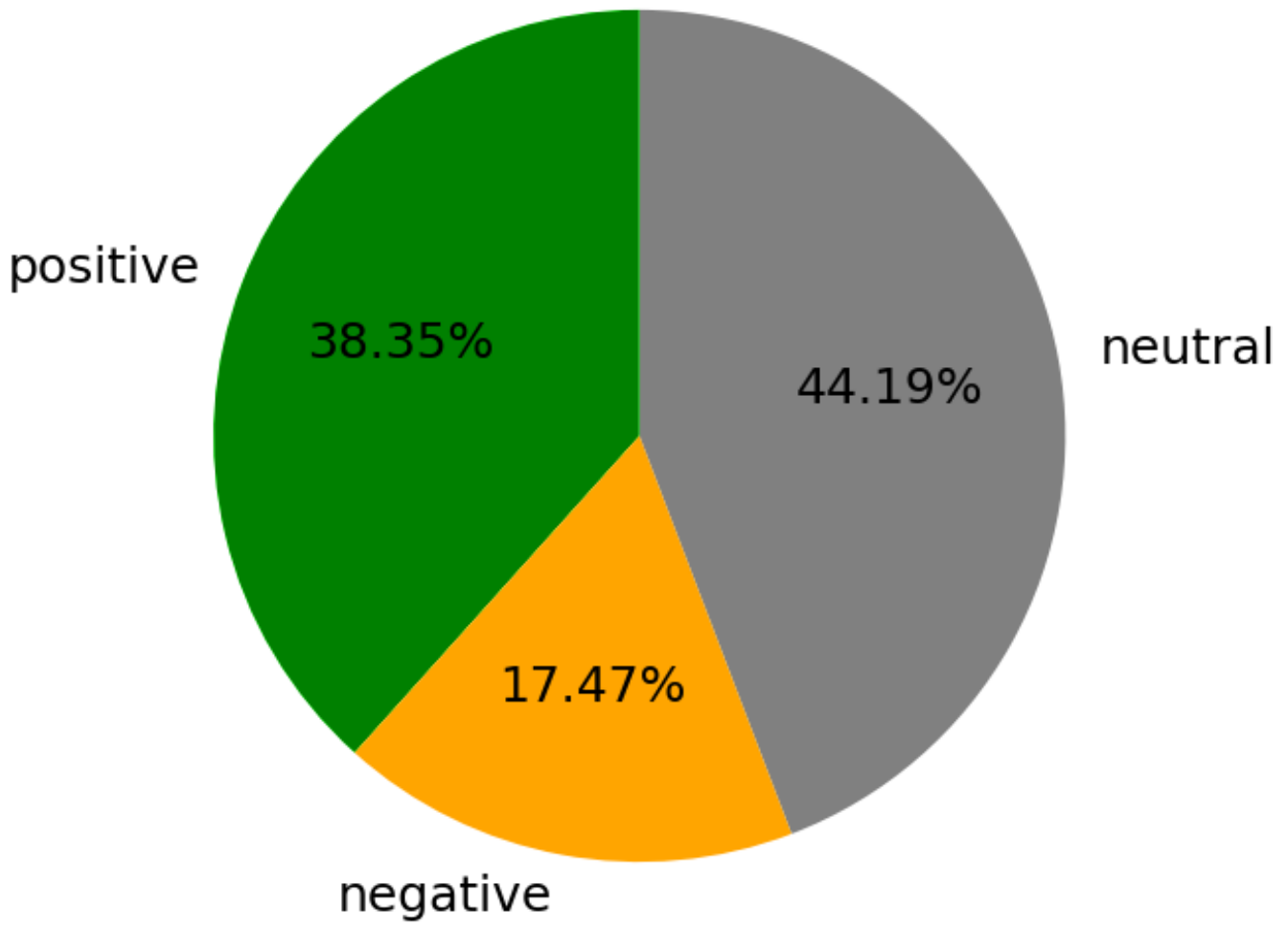

Figure 2. A pie chart that represents the variation of positive, negative, and neutral posts about COVID-19 in 2020

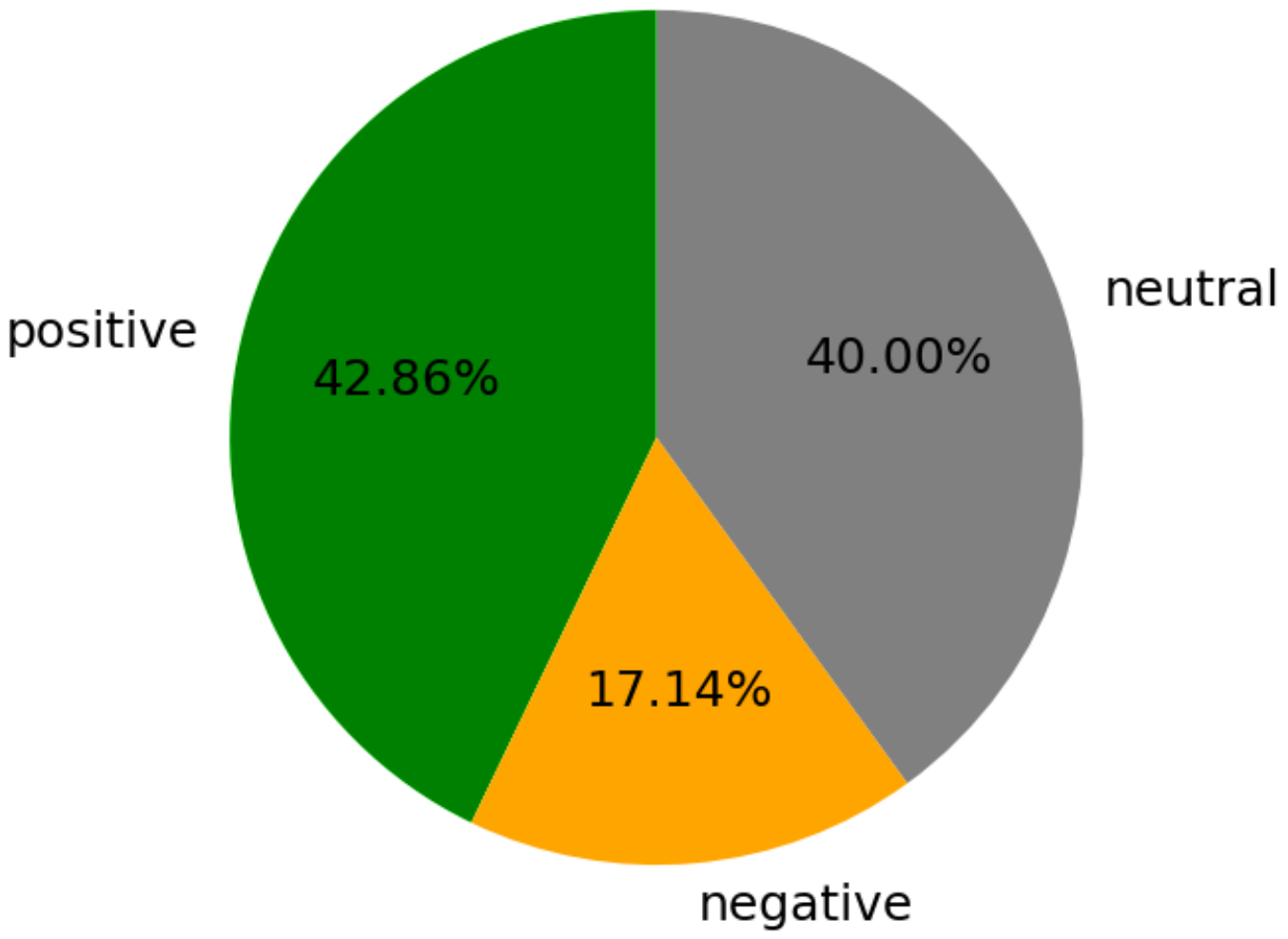

Figure 3. A pie chart that represents the variation of positive, negative, and neutral posts about COVID-19 in 2021

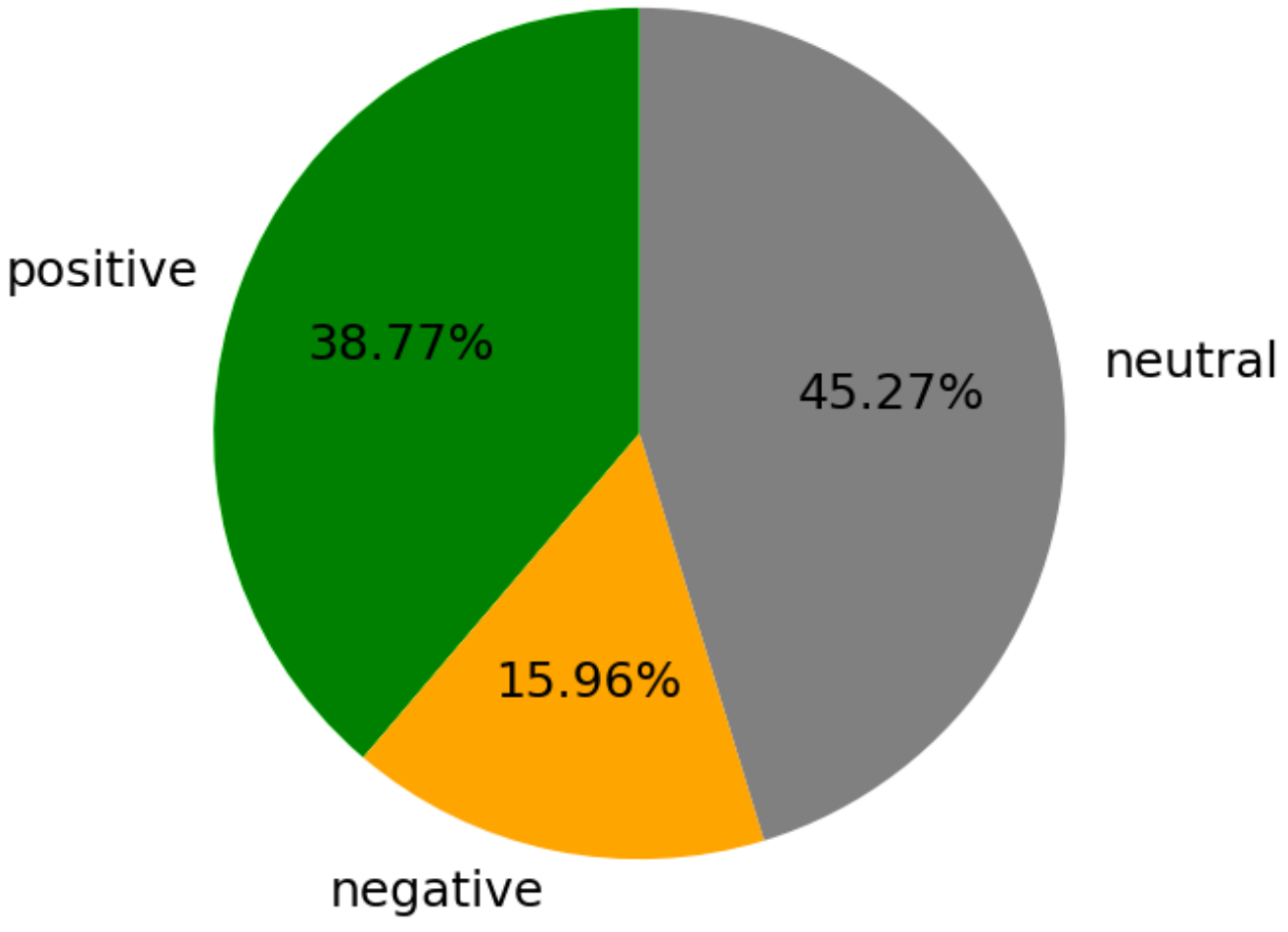

Figure 4. A pie chart that represents the variation of positive, negative, and neutral posts about COVID-19 in 2022

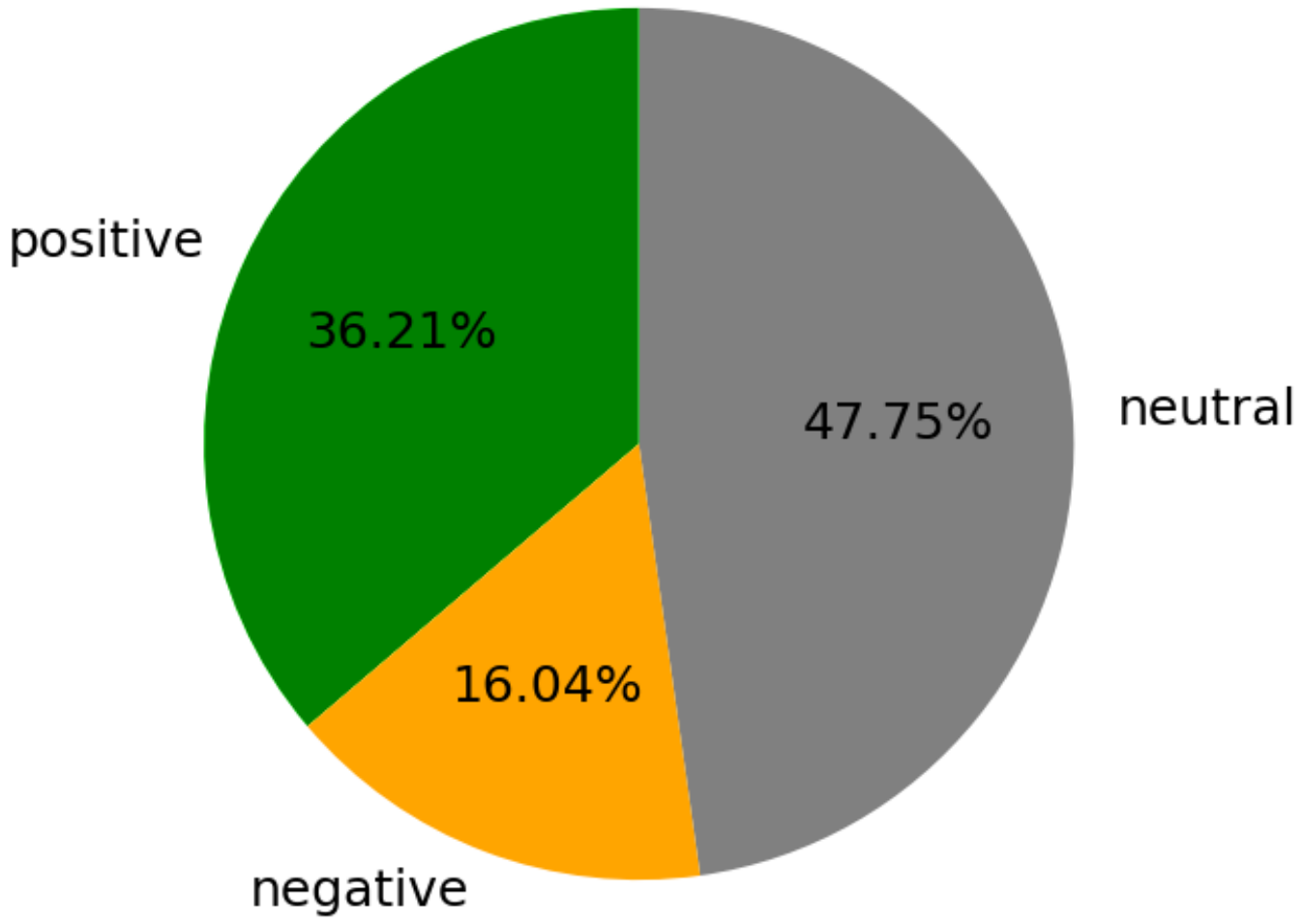

Figure 5. A pie chart that represents the variation of positive, negative, and neutral posts about COVID-19 in 2023

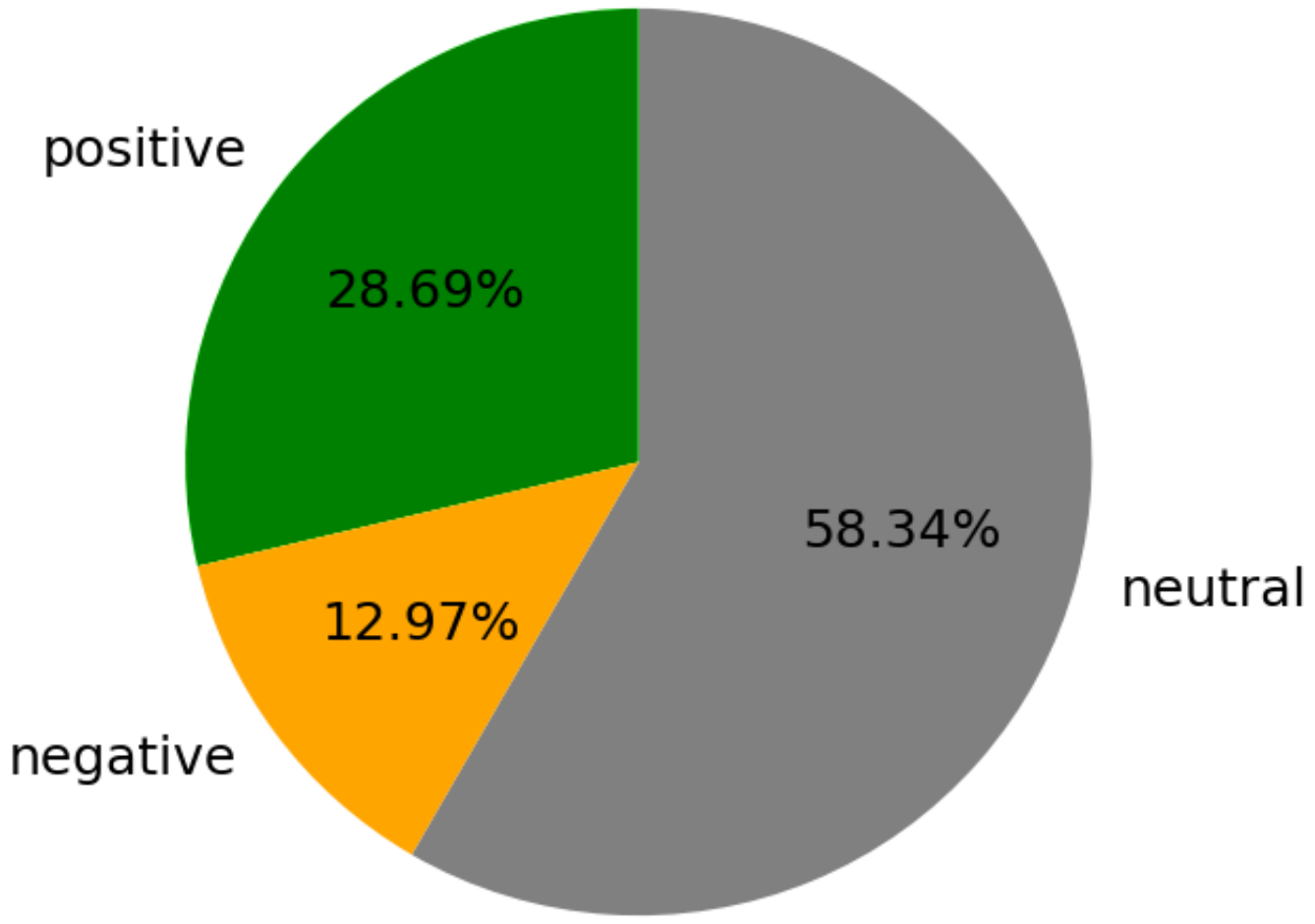

Figure 6. A pie chart that represents the variation of positive, negative, and neutral posts about COVID-19 in 2024

It is worth mentioning here that Figure 6 presents the findings of sentiment analysis for 2024 by taking into account Instagram posts about COVID-19 published between January 1, 2024, and September 23, 2024, as September 23, 2024, was the most recent date, at the time of development of this dataset. The sentiment distribution across different years provides an interesting view of how public discourse on Instagram shifted during the COVID-19 pandemic. In 2020, the first year of the global outbreak, around 38% of posts were positive and 44% were neutral. The high percentage of neutral posts probably indicates that a considerable percentage of the conversation on Instagram during the initial stages of the pandemic was centered around sharing information and updates. At the same time, the positive posts probably reflect early optimism in the face of uncertainty. Approximately 17% of the posts were classified as negative, which might indicate the challenges or anxieties some individuals experienced during the early days of the pandemic.

In 2021, the percentage of positive posts increased to 42.86%. This could be associated with developments such as the global vaccine rollout, which may have offered hope to many people. However, negative sentiment remained fairly consistent at around 17%, suggesting that concerns persisted, possibly due to emerging variants or vaccine distribution challenges. The neutral sentiment, which decreased to 40%, still suggests a significant presence of informational content in social media discussions. The analysis from 2022 shows a slight decline in positive sentiment to about 39%, with neutral sentiment increasing to roughly 45%. This shift might indicate a continued focus on factual or routine updates as people adjusted to living with the pandemic. Negative sentiment was lower at around 16%, possibly reflecting reduced public concerns as the pandemic's progression became more predictable. By 2023, positive sentiment decreased further to 36.21%, while neutral sentiment increased to nearly 48%. This trend suggests a steadying of the emotional tone, with fewer changes in public sentiment toward COVID-19. In 2024, neutral sentiment reached its peak at around 58%, while positive sentiment dropped to its lowest point at about 29%. This might suggest that discussions about the pandemic have now become more routine, with emotional engagement, whether positive or negative, being somewhat reduced. Negative sentiment also saw its lowest point, around 13%, which could suggest a diminishing sense of urgency or concern as the pandemic has now become less central to public discourse.

Next, the analysis of hashtags was performed. It was observed that there are 535,021 distinct hashtags in this dataset with the top 10 hashtags in terms of frequency being #covid19 (169865 posts), #covid (132485 posts), #coronavirus (117518 posts), #covid_19 (104069 posts), #covidtesting (95095 posts), #coronavirusupdates (75439 posts), #corona (39416 posts), #healthcare (38975 posts), #staysafe (36740 posts), and #coronavirusoutbreak (34567 posts). Figure 7 shows a word cloud of the hashtags present in this dataset. Thereafter, language-specific sentiment analysis was performed. For paucity of space, the results of performing this analysis for the top 50 languages (in terms of frequency) are presented in Table 4.

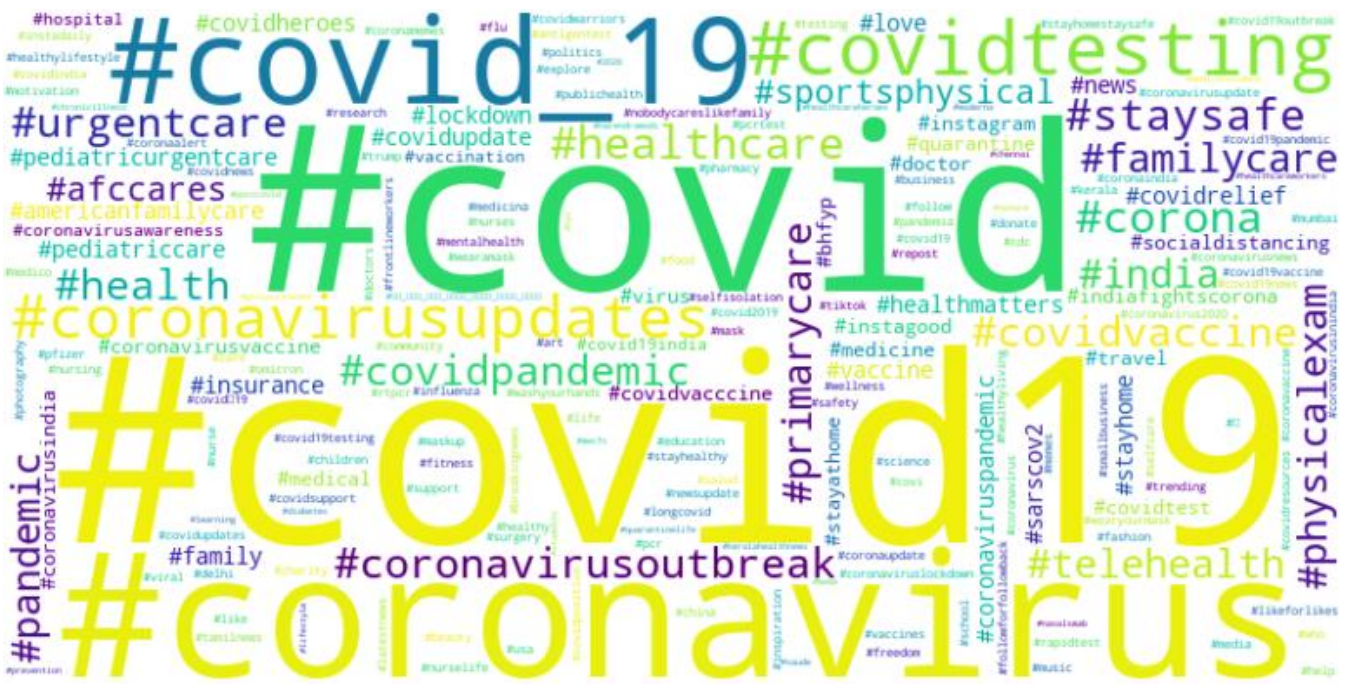

Figure 7. A word cloud analysis of the different hashtags present in this dataset

Table 4. Results of Language-specific Sentiment Analysis

| Language | Positive (%) | Negative (%) | Neutral (%) |
|---|---|---|---|
| English | 49.68 | 14.84 | 35.48 |
| Spanish | 7.01 | 8.95 | 84.04 |
| Hindi | 4.4 | 57.04 | 38.56 |
| Portuguese | 6.53 | 7.98 | 85.49 |
| Indonesian | 14.67 | 8.08 | 77.25 |
| Tamil | 0.8 | 6.04 | 93.16 |
| Arabic | 16.43 | 6.85 | 76.72 |
| German | 6.58 | 11.28 | 82.14 |
| Italian | 10.13 | 21.02 | 68.85 |
| Turkish | 17.64 | 13.26 | 69.11 |
| Marathi | 9.38 | 38.73 | 51.9 |
| Thai | 3.37 | 12.01 | 84.62 |
| Telugu | 11.14 | 12.85 | 76.01 |
| Malay | 23.15 | 12.99 | 63.86 |
| French | 17.83 | 29.32 | 52.86 |
| Gujarati | 3.45 | 31.14 | 65.41 |
| Japanese | 23.2 | 28.83 | 47.97 |
| Urdu | 7.87 | 36.79 | 55.34 |
| Malayalam | 1.67 | 11.25 | 87.07 |
| Persian | 7.88 | 21.36 | 70.76 |
| Kannada | 4.13 | 22.3 | 73.57 |
| Bengali | 7.01 | 17.71 | 75.28 |
| Chinese (Traditional) | 10.53 | 18.23 | 71.24 |
| Chinese (Simplified) | 5.01 | 46.46 | 48.53 |
| Russian | 4.55 | 5.56 | 89.89 |
| Punjabi | 0.95 | 3.05 | 96.01 |
| Dutch | 18.12 | 8.96 | 72.92 |
| Greek | 6.7 | 5.92 | 87.37 |
| Polish | 12.95 | 13.39 | 73.66 |
| Azerbaijani | 2.84 | 7.78 | 89.37 |
| Odia (Oriya) | 0.17 | 4.81 | 95.02 |
| Korean | 23.16 | 33.82 | 43.01 |
| Swedish | 13.74 | 29.86 | 56.4 |
| Sinhala | 4.11 | 7.12 | 88.77 |
| Vietnamese | 8.77 | 12.66 | 78.57 |
| Ukrainian | 4.61 | 4.96 | 90.43 |
| Bosnian | 16.46 | 12.66 | 70.89 |
| Romanian | 6.78 | 11.44 | 81.78 |
| Catalan | 13.3 | 20.6 | 66.09 |
| Filipino | 25.76 | 31 | 43.23 |
| Slovak | 9.22 | 18.45 | 72.33 |
| Nepali | 6.4 | 49.26 | 44.33 |
| Swahili | 6.4 | 31.03 | 62.56 |
| Cantonese | 5.56 | 18.69 | 75.76 |
| Czech | 13.27 | 26.02 | 60.71 |
| Danish | 17.58 | 13.19 | 69.23 |
| Croatian | 8.14 | 8.14 | 83.72 |
| Finnish | 21.3 | 33.73 | 44.97 |
| Albanian | 7.36 | 9.82 | 82.82 |
| Oromo | 11.2 | 9.6 | 79.2 |

As can be seen from Table 4, English had a balanced mix of positive (49.68%) and neutral (35.48%) posts. This may be due to the broad spectrum of content created and disseminated on Instagram in English. Spanish and Portuguese had notably high proportions of neutral sentiment, around 84% and 85%, respectively, which might indicate that discussions in these languages were more focused on sharing information than expressing strong emotional responses. In contrast, Hindi displayed a high proportion of negative sentiment at around 57%, potentially indicating greater challenges or a more emotionally charged public conversation during key moments of the pandemic in Hindi-speaking regions. Indonesian content, with around 15% positive and 77% neutral sentiment, suggests a somewhat balanced mix of hope and practical discourse. However, the large neutral component aligns with the overall trend seen in many languages, where the need for information dissemination probably dominated the tone of posts. These observations of trends in sentiments across years and languages highlight the diversity of COVID-19-related content creation and dissemination on Instagram.

In the remainder of this section, the compliance of this dataset with the FAIR principles of scientific data management [76] is explained. The FAIR principles include four key aspects of scientific data management, namely Findability, Accessibility, Interoperability, and Reusability. The components of the FAIR Principles exhibit interrelationships while maintaining autonomy and distinctiveness. The aforementioned principles delineate specific factors to be taken into account in modern data publication settings, particularly in relation to facilitating both human and computerized methods of depositing, exploring, accessing, collaborating, and reusing data. The principles may be followed in various configurations and progressively as data providers' publication settings progress towards higher levels of 'FAIRness'. Essentially, the FAIR principles endeavor to cultivate a more cooperative and transparent research landscape, facilitating the exchange of knowledge and bolstering the lasting influence of scientific investigations related to database development and database management [76]. There are several examples of datasets that comply with the FAIR principles of scientific data management such as the RCSB Data Bank [77], the Pfam protein families database [78], the Comparative Toxicogenomics Database [79], the Immune Epitope Database [80], Pdebench [81], WikiPathways [82], the open reaction database [83], MGnify [84], and MiMeDB [85]. This dataset is findable, as it is assigned a unique and permanent DOI by IEEE Dataport, making it easily discoverable for researchers across disciplines. It is accessible globally via this DOI, provided there is internet connectivity and the device used to access the internet is functional. The dataset is interoperable, as the data in this dataset is available in a standard format (.xlsx file) that can be downloaded, read, and analyzed across different computer systems, frameworks, and applications. Finally, the dataset adheres to the principle of reusability, as it can be reused any number of times for studies focused on different aspects of COVID-19-related content creation and dissemination on Instagram, such as sentiment analysis, language analysis, topic modeling, language-specific sentiment analysis, and related focus areas.

This paper has a few limitations. First, although not explicitly mentioned in the description of the twitter-xlm-roberta-base-sentiment model, it was found that the model can handle up to 945 characters. Therefore, only the first 945 characters from the preprocessed version of each Instagram post (which was not published in English) were used for sentiment analysis. Second, the Google Translate API was used for language detection. Even though the Google Translate API has a high level of accuracy, it is not 100% accurate, and these language detections were not verified by native speakers of these respective languages. Third, the findings related to sentiment analysis presented in this paper are based on this dataset. As conversations on Instagram keep evolving on a frequent basis, it is possible that if new data related to COVID-19 posts on Instagram is collected in the future and sentiment analysis is performed on the same, the results obtained from such a study may vary from the results presented in this paper. Finally, it was observed during the data mining process that a small percentage of the Instagram posts involved the usage of COVID-19-related hashtags in Instagram posts that were not related to COVID-19. This pattern of irrelevant hashtag usage by certain Instagram users was probably to gain more engagement on Instagram as COVID-19-related hashtags are usually more popular on Instagram than several other hashtags. These posts were removed by manually reviewing the dataset. As this process was performed manually, it might not have been the most optimal process for the identification and removal of such posts.

## V. CONCLUSION

This paper presents a novel dataset of more than half a million Instagram posts related to COVID-19, collected from 2020 to 2024 and spanning more than 150 languages. The dataset is labeled for sentiment analysis and is expected to serve as a valuable resource for studying public sentiment toward COVID-19 across different regions, languages, and time periods. The sentiment analysis revealed distinct changes in the proportions of positive, negative, and neutral sentiments during the pandemic, highlighting how public conversations and sentiments evolved as COVID-19 spread and continued to impact public health globally. In 2020, for instance, 38.35% of the posts expressed positive sentiment, which gradually declined to 28.69% by 2024. During the same period, neutral sentiment increased from 44.19% to 58.34%, suggesting that as the pandemic continued, more of the online discussions probably became centered on routine updates or factual information. Negative sentiment fluctuated between 12.97% and 17.47%, remaining relatively stable. These shifts demonstrate how public sentiment transformed over time, with neutral sentiment gaining prominence, possibly reflecting a shift towards more pragmatic or less emotionally charged conversations.

The results of language-specific sentiment analysis highlighted significant differences in how various communities communicated on Instagram. For example, posts in English were largely positive, with 49.68% of them expressing positive sentiment and 35.48% being neutral. This suggests that while a significant portion of the discussions conveyed optimism, a considerable amount of content was likely more informational or neutral in tone. In contrast, languages like Spanish and Portuguese were dominated by neutral sentiment (84.04% and 85.49%, respectively), potentially emphasizing the informational nature of discussions in these communities. Hindi, on the other hand, had a notably higher negative sentiment at 57.04%, reflecting the more challenging circumstances faced in regions where Hindi is widely spoken. As COVID-19 continues to influence the public discourse on social media platforms, this dataset is expected to serve as a useful resource for the investigation of a wide range of research questions. Future work would involve performing topic modeling and topic-specific sentiment analysis to identify the trends of sentiment on Instagram that are associated with different topics related to COVID-19.

## Conflicts of Interest

The author declares no conflicts of interest.